# Impossibility Results in AI: A Survey

Mario Brcic[1], Roman Yampolskiy[2]
[1]University of Zagreb Faculty of Electrical Engineering and Computing
[2]University of Louisville
mario.brcic@fer.hr, roman.yampolskiy@louisville.edu

**Abstract**
An impossibility theorem demonstrates that a particular problem or set of problems cannot be solved as described in the claim. Such theorems put limits on what is possible to do concerning artificial intelligence, especially the super-intelligent one. As such, these results serve as guidelines, reminders, and warnings to AI safety, AI policy, and governance researchers. These might enable solutions to some long-standing questions in the form of formalizing theories in the framework of constraint satisfaction without committing to one option. We strongly believe this to be the most prudent approach to long-term AI safety initiatives. In this paper, we have categorized impossibility theorems applicable to AI into five mechanism-based categories: deduction, indistinguishability, induction, tradeoffs, and intractability. We found that certain theorems are too specific or have implicit assumptions that limit application. Also, we added new results (theorems) such as *the unfairness of explainability*, the first explainability-related result in the induction category. The remaining results deal with misalignment between the clones and put a limit to the self-awareness of agents. We concluded that deductive impossibilities deny 100%-guarantees for security. In the end, we give some ideas that hold potential in explainability, controllability, value alignment, ethics, and group decision-making. They can be deepened by further investigation.

> "This, then, is the ultimate paradox of thought:
> to want to discover something that thought itself cannot think."
> S.Kierkegaard

## 1    Introduction

An impossibility theorem demonstrates that a particular problem cannot be solved as described in the claim, or that a particular set of problems cannot be solved in general. The most well-known general examples are Gödel's Incompleteness theorems [1] and Turing's undecidability results [2] in logic and computability, as well as Fermat's Last Theorem in number theory. The similar, and connected, is the notion of no-go theorems that state the physical impossibility of a particular situation. These results, though in themselves do not point to the solutions, are useful in the sense that they guide the direction for future efforts in AI in general, but in our case, the interest is in AI safety and security. For example, it may point to difficulties in verifiability for some of the current approaches [3]. In physics, there is an idea of restating the whole field in the terms of counterfactuals and what is possible and impossible in the system [4]. The authors think this will enable solutions to some long-standing questions in the form of formalizing theories in the framework of constraint satisfaction without committing to one option. A similar view regarding the utilization of the constraint satisfaction approach to many questions in philosophy is expressed by Wolpert [5]. Moreover, automated proof [6] and search [7] procedures for impossibility theorems based on constraint satisfaction were already proposed in the domain of social choice theory.

First, we shall present a classification of all relevant impossibility results based on two independent axes: *mechanism* and *domain*. Mechanism axis should help in finding new impossibility results. Under that axis, all the impossibility theorems are shown to be neatly subsumed under the problems with capacity disparity where several related objects differ in size. Domain axis is informative for applications in system engineering which combines symbolic and learned modules. Then, we shall present the current impossibility results that we find relevant for AI. That includes work made specifically in AI, but also work in other fields such as mathematics, physics, economy, social choice theory, etc. In the process, we shall show new results. The first result states the unfairness of explainability. It is the first induction impossibility theorem pertaining to explainability as all the previous ones were addressing the perspective of deduction. The second result deals with misaligned embodiments when the cloning procedure can produce an adversarial situation. The third result states the impossibility of being perfectly self-aware.

Previous works cover similar topics within the scope of AI safety [8]–[11], but none focused on impossibility theorems as a family, their utility, structure, and connections to other fields. Especially in the light of recent advances of language models on math [12] and programming [13] that have spurred speculations, we strongly believe that research in AI safety should start at hypothesized limitations of AI and work inwards to the current technology. This approach is proactive and more robust than the reactive approach that plays catch-up with current results.

The rest of the paper is organized as follows. In section 2 we introduce basic definitions. Section 3 contains the relevant work presented under newly defined classification. In section 11 we focus on impossibility theorems developed in the field of AI safety. The discussion is offered in section 17, and ideas for future research are listed in section 18. Finally, we conclude the article in section 18.

## 2     Basic definitions

We shall not impose strict formalization, but we shall keep at the level of lawyerese in this paper to ensure wide readability of material. Other papers will cover more formalized arguments. It is evident that our investigation is done from the perspective of assumption that *intelligent behavior that can achieve its goals is computable*.

***System*** *is any non-empty part of the universe.*

***State*** *is the condition of the universe.*

***Control*** *of system A over system B means that A can influence system B to achieve A's desired subset of state space.*

Usually, with control, we aim at output controllability (from control theory). Such control is not sufficient for safety – as we often make unsafe choices ourselves. Different modes of "influence" and "desire" are possible. With regards to that, Yampolskiy in [8] mentions four types of control: explicit (strict), implicit, aligned, and delegated. **Explicit control** agent takes expressed desires literally and acts on them. **Implicit control** agent uses common sense as a safety layer over explicit control to slightly reformulate the expressed desire and acts on it. **Aligned control** agent adds intention inference over implicit control in order to postulate the intended desire and acts on it. **Delegated control** agent decides for itself the subject's desire that is long-term-best for the subject and acts on it.

***Intelligence*** *is the ability for an information processing system to adapt to its environment with insufficient knowledge and resources. [14].*

***Safety*** *of system A is the property of avoidance of going out of A's desired subset of the state space.*

Safety is pressed with finding the worst-case guarantees – which is modeled as adversarial games that assume the ideal adversary.

***Stability*** *of state S for system A is the intrinsic tendency to return to A's desired subset of state-space after being perturbed.*

***Robustness*** *of state S for system A is the property of staying within A's desired subset of the state space despite perturbations.*

***Catastrophic outcome*** *for system A is any state from which the return to A's desired subset of state space is impossible.*

***Alignment*** *within the ensemble of systems $A_1...A_n$ is the property that each system $A_i$ achieves greater than or equal benefit from working together than if any subset of agents acting self-interestedly.*

Alignment is about finding values that would make the game cooperative in a long term. There are plenty of open questions regarding the topic of alignment. How to achieve cooperativity in game over long periods? Is the game allowed to temporarily deviate from perfect cooperation? The problem for humans is that we do not have consistent short- and long-term values. Sometimes we have to suffer in short term (like in sports) in order to prosper in the long term. How to define alignment with such preferences?

## 3  Impossibility theorems

Impossibility theorems boil down to some contradiction. The potential to find impossibility theorems lurks at the appearance of paradoxes. Paradoxes are simple implications that there is some constraint, limit, we were not aware of when we unknowingly reached out of the feasible area and reaching some contradiction in our stated goals. Finding impossibility theorems circumscribes our knowledge of possible which enables us to direct our efforts better and do better risk management. These can help even when we do not know the status of final solutions, such as for AGI and AI Safety, in a form of directing constraint satisfaction search – both in formalizing ideas and committing to certain hypotheses. It might be that this is the most prudent approach for approaching hard problems with long horizons to finding solutions. It works on a meta-level of scientific investigation by upending the way research is done, in the process creating results about the problem even when oblivious about the actual solution.

We propose two independent axes in classification of impossibility results:
  I. **mechanism** axis which is related to the capacity disparity
  II. **domain** axis which pertains to the characteristics of systems where they appear:
   a) *symbolic (S)* – in formal mathematical models
   b) *data (D)* – in empirical systems that learn about the world from observations

## 3.1  Mechanism

Impossibility theorems pertinent to AI are mostly related **to the problem of capacity disparity**. Namely, intuitively we operate between the domains that differ in size. For example, when expanding from intrinsically smaller to a greater domain, or in the opposite direction when contracting from bigger to the smaller domain. We have organized all the impossibility theorems into 5 subcategories (Figure 1):

1. Deduction (D),
2. Induction (I),
3. Indistinguishability (N),
4. Tradeoffs (T), and
5. Intractability (P).

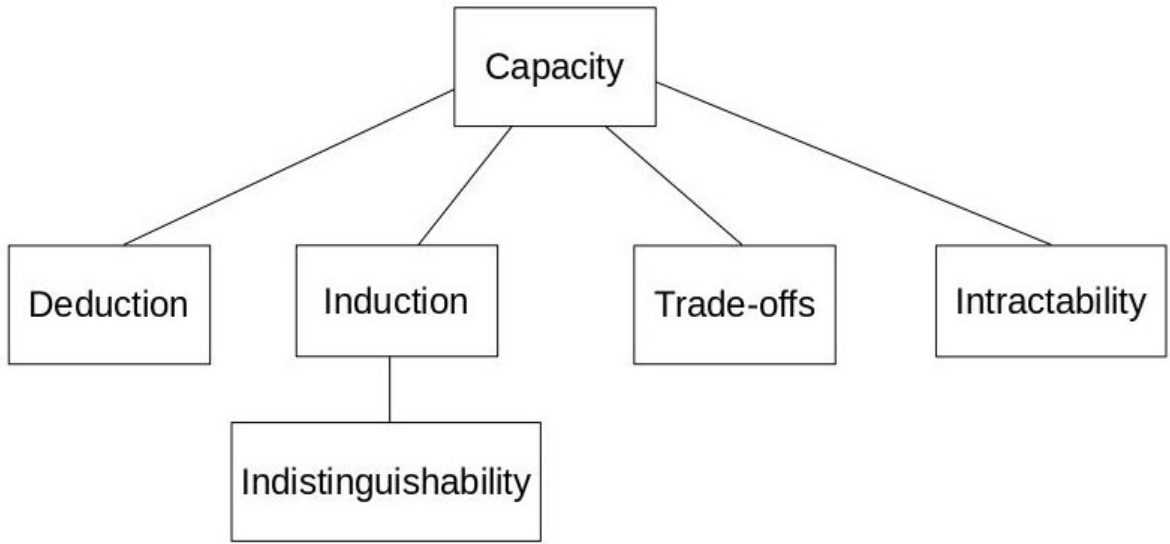

*Figure 1 Proposed mechanism-based categorization of impossibility theorems in AI*

For example, **deduction** tries to go beyond its size capacity by going from countable to uncountable infinite (Turing's computability), in Gödel's terms by going from provable to true statements, or in Chaitin's terms from lower to greater Kolmogorov complexity of formal systems. Self-referential paradoxes go beyond these capacity limits by including itself in its own definition and negating itself. From that follows an infinite fractal-like growth where we used finite means to express infinite without a fixed-point. Examples include unverifiability. The proofs in this category often use Lawvere's theorem [15], [16] in the disguise of Cantor's diagonalization and liar's paradox.

In terms of **induction**, we have to find a model within a (possibly infinite) set of plausible models based on finite dataset/experience. Hence, we have the finite capacity of experience to guide our search within a set having multitude (possibly infinitude) of elements. There are too many inductive inferences that can be made. Induction as a general operation is prone to the problems emanating from Hume's problem of induction and Goodman's new riddle of induction [17]. No Free Lunch theorems [18], [19] deal with induction and are a formalization of Goodman's new riddle of induction [20]. As such, they can be the basis of a vast number of induction-based impossibilities. Examples include unpredictability.

Also, in special cases of induction, we have a problem of disentangling, ie. indistinguishability (**non-identifiability** and **unobservability**). We wish to get the inner structure from the limited entangled data. No amount of data can enable identification in the cases of non-injective transformations that produced the data. This means the impossibility of learning even in the limit. There is an inevitable **loss of information** going from the origin through the (capacity reducing) transformation. That prohibits the recovery of full information and leads to observational equivalence. Examples include unobservability in control theory [21].

**Trade-offs** are inherent to multicriteria decision making, or "you can't have it all". The problem of size capacity is evident here in the inability to obtain the point with the individual maximal value of each component at the Pareto front (and hence full hypervolume indicator [22]).

In **intractability,** we have physical limits on capacity (in memory, computing power,…) which prohibits efficiently reaching solutions. This is not only a set-theoretical, computation-independent limit, but it also enables relative comparisons, for example through Karp's hierarchy [23].

We have listed in Table 1, below, impossibility theorems we deemed important for for field of AI. There are many more impossibility theorems in mathematics and physics, but they are not (to us) evidently related to AI. There are also many impossibility theorems in machine learning, but we omitted them due to the too-narrow scope.

*Table 1. Impossibility theorems of interest to AI researchers. Mechanism categories: D – deduction, I – induction, N - indistinguishability , T – tradeoffs, P – intractability. Domain categories: S – symbolic, D - data.*

| Name | Source | Proven (Y – yes, N – no,  ~ - not rigorous) | Mechanism category (D;I;N;T;P) | Domain category (S; D) |
|---|---|---|---|---|
| Unobservability | [21] | Y | N | D |
| Uncontrollability of dynamical systems | [21], [24] | Y | N | D |
| Good Regulator Theorem | [25] | Y | N | D |
| Law of Requisite Variety | [26] | Y under perfect information and infinite speed | N | D |
| Information-theoretical control limits | [27] | Y | N | D |
| (Anti)codifiability thesis | [28]–[30] | N | I | D |
| Arrows impossibility theorem | [31] | Y | T | S |
| Impossibility theorems in population ethics | [32] | Y | T | S |
| Impossibility theorems in AI alignment | [33] | Y | T | S |
| Fairness impossibility theorem | [34], [35] | Y | T | S |
| Limits on preference deduction | [36] | Y | N | D |
| Rice's Theorem | [37] | Y | D | S |
| Unprovability | [1] | Y | D | S |
| Undecidability | [2], [38] | Y | D | S |
| Chaitin Incompleteness | [39] | Y | D | S |
| Undefinability | [40] | Y | D | S |
| Unsurveyability | [41] | N | P | D |
| Unlearnability | [42], [43] | Y | D | D |
| | [44] | Y | P | |
| Unpredictability of rational agents | [45], [46] | Y | I | D |
| No Free Lunch - supervised learning | [18] | Y | I | D |
| No Free Lunch - optimization | [19] | Y | I | D |
| Free Lunch in continuous spaces and coevolutionary | [47], [48] | Y | I | D |
| Unidentifiability | [49]–[52] | Y | N | D |
| Physical limits on inference | [5], [53]–[55] | Y | D+I | D |
| Uncontainability | [56] | Y | D | S+D |
| Uninterruptibility | [57]–[61] | N only under limited assumptions - opened | D | D |
| Löb's Theorem (unverifiability) | [62] | Y | D | S |
| Unpredictability of superhuman AI | [63], [64] | Y~ definition of superhuman? | I | D |
| Unexplainability | [65] | N based on explanation=proof and using at least one of: (unprovability, undefinability), assuming honesty or full model | D | D |
| Incomprehensibility | [65] | N based on explanation=proof and using at least one of: (unverifiability, unsurveyability, undefinability), assuming honesty or full model | D | D |
| | [66] | Y comprehension=producing proof and halting games | | |
| k-incomprehensibility | [67] | N, just definitions | I | D |

| | | | | | |
|---|---|---|---|---|---|
| Unverifiability | [68] | Y | D | S | |
| Unverifiability of robot ethics | [69] | Y | D | S | |
| Intractability of bottom-up ethics | [70], [71] | Y | P | D | |
| No-flattening theorems for deep learning | [72] | Y | P | D | |
| Efficiency of computing Boolean functions for multilayered perceptrons | [73] | Y | P | D | |
| Goodheart's Law (Strathern) | [74], [75] | N | I | D | |
| Campbell's law | [76] | N | I | D | |
| Reward corruption unsolvability | [77] | Y | I | D | |
| Uncontrollability of AI | [8] | Y~ under degenerate conditions | D | D | |
| Impossibility of unambigous communication | [78] | Y under strict assumptions | I | D | |
| Unfairness of explainability | here | Y~ proof sketch | I | D | |
| Misaligned embodiment | here | Y~ proof sketch | T | D | |
| Limited self-awareness | here | Y~ proof sketch | I | D | |

### 3.1.1 Deduction Impossibility Theorems

Limits of deduction are limits on our capability to achieve perfect certainty in facts. The vast majority of listed results here use Lawvere's fixed-point theorem [15], [16] as the basis of proofs, i.e. more specifically a combination of Cantor's diagonalization and liar's paradox.

The most basic results here are Gödel's incompleteness theorems [1] addressing unprovability and Turing's work [2] covering undecidability. In addition to the aforementioned which cover the processing, Gregory Chaitin provided additional incompleteness results [39] that cover input sizes measured by Kolmogorov's complexity. Chaitin's incompleteness theorem states the existence of a limit on any formal system to prove Kolmogorov complexity of strings beyond some length. There are even conjectures that information complexity might be the source of incompleteness [39], [79] whereby *the theorems of finitely stated theories cannot be significantly more complex than the theory itself*.

**Rice's theorem** [37] is a generalization of Turing's undecidability of the halting property of programs to any sufficiently complex property. This makes it an ideal tool for finding and proving deduction limitations in AI, within its assumptions. Löb's theorem [62], informally put, states that a formal consistent system cannot, in general, prove its own soundness.

Regarding formal semantics, Tarski's undefinability theorem [40] states that truth in a formal system cannot be defined within that system. Measuring semantic information yield by deduction is also poised by paradoxes. Bar-Hilel-Carnap paradox [80] in classical semantic information theory entails that contradiction conveys maximal information. Hintikka's scandal of deduction [81] points to the fact that the information yield of truthful sentences is zero since their information is already contained in the premises.

There is vast work in impossibility theorems in beliefs which is an extension of Gödel's work. Huynh and Szentes [82] have demonstrated irreconcilability between two notions of self-belief. In [83], [84] the paradox of self-reference was extended to the games with 2 or more players which yield impossible beliefs.

Deductive impossibility results have been found in machine learning as well regarding the properties of learning algorithms and processes given the data. Authors in [42], [43] have shown that learnability can be undecidable and unprovable for certain problems. Finding if an encodable learning algorithm always underfits a dataset is undecidable [85], even with unlimited training time.

Inference devices covered in the series of papers by Wolpert [5], [53], [54] are an extension from pure deductive systems to the general notion of inference devices which covers deduction, induction (prediction, retrodiction), observation, and control while the device itself is embedded in a physical universe. Additional limitations are found, both in a logical and stochastic sense. For example, limits are found on strong and weak inference, control, self-control, and mutual control between the two distinguishable devices. The limits are also put on prediction, retrodiction, observation, and knowledge. This work was extended in [55] to find the constraints on modeling systems from within systems using the frame of relativity. The concluding impossibility theorem states that the universe cannot be completely modeled from within the universe.

### 3.1.2 Indistinguishability Impossibility Theorems

**Observability** is the ability to infer the state of a black-box system from its input/output data. **Identifiability** (ie. parameter and/or structural identifiability) is a special case of observability for constant elements of the system whereby we only need to infer the values of those constant elements. There are limits to both observability and identifiability, and the limits are caused by non-injective mappings [52] which inevitably lose information.

Identifiability and observability are important for the control over systems.
**Controllability** in control theory can be of two kinds: state and output. **State controllability** is the ability to control the inner state of the system. **Output controllability** is the ability to control the output of the system. State uncontrollability is the dual of unobservability. That is, state controllability is impossible without observability [21], [24].

**Good regulator theorem** [25] relates output controllability and identifiability (modeling), but only in a sense of optimal control, not sufficient control. It states that maximally simple among *optimal* regulators must behave as an image of the controlled system under a homomorphism. Sufficiently good regulators need not be optimal and the generalization of such theorem would be interesting.

**Law of requisite variety** [26] states that variety in outputs can only be reduced by the state complexity of the controlling mechanism. **Information theory state-control limit** [27] says that only up to information observed from the system can be used to reduce the entropy of the system.

In the absence of additional biases, general nonlinear independent component analysis has an infinity of solutions that are indistinguishable [49]. Similar is shown for unsupervised learning of disentangled representations [51]. There are also well-known non-identifiability limits to causal discovery from the data [50].

### 3.1.3 Induction Impossibility Theorems

Limits on induction constrain our ability to infer latent factors. Here we will ignore the problem of indistinguishability by just looking at equivalence classes of models indistinguishable in the limit. In this case, there is a possibility to learn the true equivalence class asymptotically. But, given some prefix of experience, there may be a multitude of candidate classes. This is pointed out in Hume's problem of induction and Goodman's new riddle of induction [17]. **No free lunch theorems (NFL)** by Wolpert [18], [19] are the basic building blocks underlying the formalization of these limitations. They were first formulated in general supervised learning and optimization, which were subsequently unified through that framework. No free lunch theorems state that under uniform distribution over induction problems, all induction algorithms perform equally [86]. At the heart of NFL formalization is the independence of (search/learning) algorithm performance from the uncertain knowledge of the true problem at hand. That independence is materialized in the inner product formula of those two in describing the probability of attaining a performance value over the unknown problem. There are, however, free lunches if more structure is imposed on the problem, i.e. "there is no learning without bias, there is no learning without knowledge" [87]. For example, there are free lunches in continuous spaces [47] and in coevolutionary problems [48].

Goodhart-Strathern's law [74], [75] and Campbell's law [76] deal with the difficulties and the inability in finding expressible proxy numerical measures for success that are **well aligned** with inexpressible/unknown-explicitly experiential measure of success. A similar sentiment is expressed in [88] where a more detailed explanation is given for the observed difficulties, all stemming from the unpredictability of solution routes to hard or even unknown problems. Metric is a model of an imagined success, but shallow and not with perfect alignment.

**In games** with uncertainty in opponent's payoffs, it is impossible to predict the behavior of perfectly rational agents due to the feedback loop emanating from their own decisions which influence opponent's behavior [45]. Placing further restrictions on the assumptions can regain predictability. In economic situations, further limits relating to rationality, predictability and control were proved in [46]. Therein, (i) logical limits were set to forecasting the future, (ii) non-convergence of Bayesian forecasting in infinite-dimensional space was demonstrated, and (iii) impossibility of computer perfectly forecasting economy if agents know its forecasting program. These results are related to the already mentioned results in deduction-related ITs for Wolpert's inference devices and regarding beliefs in games.

Anticodifiability thesis [28]–[30] is a conjecture in moral philosophy that states that universal morality cannot be codified in a way that would be aligned in all circumstances with our inexpressible/unknown-explicitly experiential moral intuition.

### 3.1.4 Tradeoffs Impossibility Theorems

Trade-off limits constrain our attempts to achieve perfect outcomes. Examples include impossibility theorems in clustering [89], fairness [34], [35], and social choice theory (SCT) [31]. In many situations, we have to choose with respect to multiple criteria simultaneously. Often, it is the case that there is no ideal point that simultaneously optimizes all the criteria, that is achieves maximal possible hypervolume indicator [22].

In social choice theory, there are results such as Arrow's impossibility theorem [31] which states there must be a trade-off that forces choosing only a strict subset of desirable properties in voting mechanisms. In moral theory, there are different problems regarding population ethics [32] where all total orderings entail some problematic properties that contradict our intuitions. Solutions have been proposed for automated systems that search for impossibility theorems in SCT regarding rankings of objects [7].

### 3.1.5 Intractability Impossibility Theorems

Intractability limits divide possibility-in-principle and practically impossible due to the resource limitations. There are three types of intractability ITs: asymptotic, physics-based, and human-centered.

**Asymptotic intractabilities** fall neatly under the complexity theory [23], [90]. That research field is simply too rich to expand on it here. We shall only highlight the probably approximately correct (PAC) learning framework [44] by Valiant that defines the border between efficient (polynomial time) and inefficient learnability. Of our interest is also the intractability of bottom-up ethics [70], [71] which stems from the game-theoretic nature of ethics. Lin et al. in [72] have dealt with possible reasons and situations when deep learning works efficiently and when does it necessitate an exponential number of parameters in the number of variables. They prove various "no-flattening theorems" that show loss of efficiency when approximating deep networks with shallow ones. Calude et al. have extended the latter work in [73] by working on sensitive and robust Boolean functions, whereby especially parity function poses exponential difficulties for learning by multilayered perceptrons with *single* hidden layer.

**Physics-based limits** put bounds on physically implementable computation and intelligence. No-go theorems state constraints for certain implementation approaches. Currently, these limits (e.g. [91]) are quite loose and/or specific so we do not go into their details. One exception is the work of Wolpert we have previously mentioned [5], [53], [54]. That research is quite general and is an extension of previous mathematical results by embedding computational agent into the universe within which it utilizes resources for computation.

There is an area of **human-centered limits** which does not seem to be well researched and measured. Humans, as agents of finite capabilities, have strict limits with regard to explainability, comprehensibility [11], and all other aspects. One of the commonly mentioned impossibilities is unsurveyability [41] in the context of mathematical proofs.

## 3.2 Domain

Domain axis in the categorization pertains to the characteristics of systems where impossibility results appear:

1. symbolic (S) – in formal mathematical models
2. data (D) – in empirical systems that learn about the world from observations

This reflects the split in AI between model-based (i.e. symbolic/logical/deductive) and empirical data-based systems (i.e. statistical/inductive). It also covers the split between mathematics and physics in their usage of the word "theory". In the former, it is a robust and self-sufficient set of crafted principles for which we can have some understanding and rigorous analysis. In the latter, the theory is contingent and not robust to data variations as future data can falsify theories. The symbolic category is mostly hit by impossibilities due to undecidability, high complexity, and constraints. Data category, in addition to undecidability, constraints, and complexity, is subject to underdetermination and intractability.

# 4 Impossibility theorems developed in AI safety

**Uncontainability** [56] states the inability of preventing superintelligence harming people if it chooses to, by recognizing the intent ahead of time. This is due to the undecidability of harmful properties in complex programs (corollary of Rice's theorem).

**Unverifiability** [68] states fundamental limitation (or inability) on verification of mathematical proofs, of computer software, of the behavior of intelligent agents, and of all formal systems. This is a corollary of Rice's theorem as well. An extension of Rice's theorem to robot programs was proven in [69] to show impossibility of online verification of robot's ethical and legal behavior.

**Uninterruptibility** [57]–[61] states that under certain conditions it is impossible to turn off (interrupt) intelligent agent. Possibilities and impossibilities have been shown under specific assumptions and conditions.

**Unpredictability** [63], [64] states our inability to *precisely and consistently* predict what specific actions an intelligent system will take to achieve its objectives, even if we know the terminal goals of the system. The proof depends on the implicit, but the unstated definition of unaligned superhuman intelligence and forms contradiction. The form of the proof does not limit occasional imperfect but sufficiently precise predictions. The question is, short of perfection, how much predictability is sufficient?

**Unexplainability** [65] states the impossibility of providing an explanation for certain decisions made by an intelligent system that is both 100% accurate and comprehensible. Here, the explanation is taken to be a proof which is then prone to the deduction ITs such as unprovability and undefinability. What is not covered is with respect to what is accuracy measured against and does not cover truthfulness of explanation in the case of incomplete information. Explaining yourself truthfully and correctly would imply self-comprehension which is a problematic notion itself as disproved in [66].

**Incomprehensibility** [65] states the impossibility of complete understanding of any 100%-accurate explanation for certain decisions of an intelligent system by any human. It is the dual of explainability and again it is assumed that explanation is proof which leads to the use of deductive ITs. Understanding is vaguely defined as proof-checking and it is not defined how accuracy is measured. In a similar line of work Charlesworth [66] defines comprehension of some systems as the capability of producing correct proofs by fallible agents about those systems. He takes a program as a starting point, implicitly assuming its truthfulness. He then produces relations of comprehensibility and rules out self-comprehension.

**Uncontrollability** [8] states that humanity cannot remain safely in control while benefiting from a superior form of intelligence. The proof uses a Gödel-like structure that shows the impossibility of perfect control in degenerate conditions which invoke self-referential paradoxes with controls. The form of control shown to be impossible was explicit control. In fact, with such proof, *uncontrollability holds for any sufficiently complex agent over which explicit control is attempted*, including humans. Moreover, *the proof holds also for the case of attempted self-control*. This counter-intuitive notion points into the direction that more research is necessary into formalization and disentangling of the structure and assumptions of explicit control. Advanced forms and notions of control should at least resolve the status of control over oneself. More research is necessary into the status of controllability for other forms of control (implicit, aligned, delegated).

**Limits on utility-based value alignment** [33] state a number of impossibility theorems on multi-agent alignment due to competing utilitarian objectives. This is not just an AI-related topic. The most famous example is Arrow's Impossibility Theorem from social choice theory, which shows there is no satisfactory way to compute society's preference ordering via an election in which members of society vote with their individual preference orderings.

**Limits on preference deduction** [36] state that even Occam's razor is insufficient to decompose observations of behavior into the preferences and planning algorithm. Assumptions, in addition to the data, are necessary for disambiguation between the preferences and planning algorithm. This is due to non-injective mapping induced by preferences and planning algorithm that produce behavior.

**Unsolvability of reward corruption** [77] states that without simplifying assumptions it is impossible to solve reward corruption problems such as wireheading, sensory error, reward misspecification, and error in preference deduction. The proof is done via an NFL route and holds for reinforcement learning, for example. The problem can be averted under some simplifying assumptions and sufficient reward crosschecking. Otherwise, quantilisation [92] may provide more robustness.

**Impossibility of unambiguous communication** [78] denies perfectly unambiguous communication using natural language. Many examples are given to show different levels of ambiguity: phonology, syntax, semantics, and pragmatics along with contemporary NLP and AI approaches to handling them. These areas are taken together to show, under simplified assumptions, that ambiguity is inevitable in communication using natural language. Generalization of Goodhart's law for problem detection tools in AI systems is given, without proof. The intention and mechanism behind the proposed law seem to be adversarial learning.

The following subsection introduces a new impossibility result.

## 4.1 Unfairness of explainability

Let us examine the explainability process which consists of decision making, explanation generation, and verifying decision through the explanation. Let us assume that verification decides if the decision will be accepted (i.e. veto capability). If we assign subjects to those phases, we get the scheme from Figure 2 with decision maker (DM), explainer (E), and verifier (V). It is, or it will become, evident that the explainability process is an approximation to the containment process [56], whereby we bypass deductive limits only to hit the inductive limits.

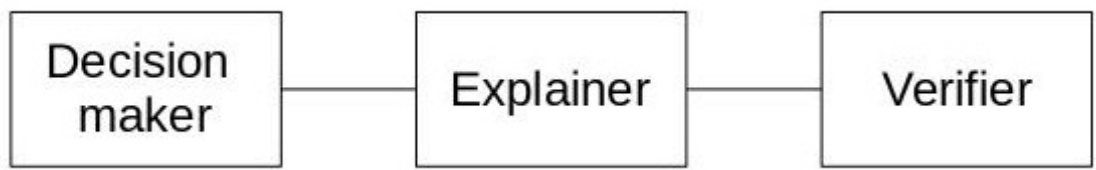

*Figure 2 Explainability process*

We can make connections to the previous work on explainability and comprehensibility using the above scheme. Explaining takes place in E with the resulting output explanation A, while comprehending takes place in V with resulting comprehension C (in this case it is a classification).

In [65], DM's decision D is assumed to be conjecture, A is assumed to be proof. From that follow the problems with producing proofs (explaining) on the side of E and problems with verifying proofs (comprehending) on the side of V. In [66], E is assumed to be the software code itself, and comprehending is producing proof with C being justifiable halting prediction.

The only perfectly correct explanation for the decision-maker's (DM) decision is the whole computational process that made the decision from all the available inputs – that is, DM execution trace (or equivalent). The principle behind that is "I did it, because of how I did it". An **explainability process** is necessary because the verifier (V) is not computationally capable enough to understand and verify DM's process. The task of the explainer (E) is to simplify explanations to be simpler than the DM's computational process. However, short of the DM's execution trace (or its equivalents), there are multiple models of lesser complexity consistent with the observed output (decision). E chooses an explanation out of the ensemble of consistent ones and thus influences the verification outcome. In some situations, the explanation might imply the benevolence of the decision, such that V decides positively. In other cases, the explanation might point out malign or erroneous decision and V decides negatively.

If these processes are distributed across different agents, we get a game where different agents might not be aligned. The interesting situation is where DM is trying to make V believe the decision is benevolent while in fact, it is not. We will look at:
- self-explaining decision-making (DM+E) - where DM and E are integrated into one agent (e.g. [93]),
- inferred explanation – where DM is separate from E.

Let us assume a setting where DM/DM+E is unaligned with V, i.e. assume dishonesty of DM. This can be reflected in the fact that DM receives greater rewards when it "tricks" V, i.e. when D is malign and C is positive. We are interested in the size of rewards DM can gain while "tricking" V. We can call such rewards **malignant gain** $f(D, C)$.

Self-explaining agent DM+E aims to maximize function $f(D, C=V(A))$ with respect to generated pair (decision D, explanation A) wherein it is constrained by the model of V. For inferred explanation DM wants to maximize reward function $f(D, C=V(A=E(DM, D)))$ with respect only to the decision D that it has under its control. In the latter case, E and V make up potential obstacles for DM.

**Theorem 1. (Unfairness of explainability)** There is an inherent strategic inequality in explainability:
a) Short of the DM execution trace (or the equivalent), it is impossible for V to differentiate between consistent(plausible) explanations of decisions.
b) DM's malignant gain increases with better models of f, E, and V.

*Proof.* *Assume that E does the simplification i.e. reduction in complexity with respect to perfectly correct explanation. For each set of outputs, there is a multitude of consistent (with some specification of consistent that can be more flexible or strict) models that could have produced that prefix. There is an indistinguishability issue for V without some further assumptions (it is in the NFL regime). This enables DM to make decisions for which there may be plausible benevolent explanations, but in fact are malign (i.e. through information hiding, greater cognitive capabilities, etc.). If DM is a black-box optimizer without any information of E, V, and f, and they may be arbitrary, DM is in the NFL regime, all policies have similar average performance. However, assuming DM was created by V(or aligned entity), certain biases and information have been instilled into DM that enable it to solve certain tasks for which it is being employed.*
*In the case of self-explanations (DM+E), the agent makes pairs (decision, explanation). The malignant gain is limited only by the problem, willingness, quality of DM+E's model of V, knowledge*

*of its preferences, and computational capabilities of DM+E to optimize for that pair, treating C as an uncertain element.*

*In the case of an inferred explanation, DM only makes a decision, and the explanation is inferred by E. Here one more layer of complexity is added since DM must have a good model of E. DM can optimize decisions, treating both A and C as uncertain elements. In the most extreme case of having perfect models, DM would be limited only by the problem and optimization capabilities.*

We can see that constructing DMs to solve certain problems and give personalized explanations gives information that can be also utilized for malignant gain. The examples of these situations do not need the presence of artificial intelligence agents but are already present in normal life: judicial processes, politics, etc.

The complexity of comprehensible explanations is limited by V's cognitive and situational capabilities. We postulate that the bigger gap between the complexity of explanation and the complexity of the real process that produced the decisions, more malignant decisions can be plausibly justified using observed history.

## 4.2 Misaligned embodiment

This section is clearly inspired by Parfit's work [94].

Let us imagine a situation where a computational agent goes through operational cloning by mistake, in one of two ways: copy or separation. A pure copy would correspond to the network transfer of the agent to another location, mistakenly not deleting the agent at the source location. Separation would correspond to the accidental loss of connectivity between the agent's coordinated components each of which has an identical copy of code, in which case they would become two separate agents.

In that situation, we shall refer to the agent's code as a **type** and an instance of that code in each agent as a **token**. We are concerned with the characteristics of types that lead to misalignment between the mistakenly operationally created clones. The decisive assumption is the existence of the "reward mechanism" with which agent's tendencies can faithfully be algorithmically modeled. Especially interesting is the locus of self in self-modeling through reward. Additionally, we assume that the improvement of reward utilizes some scarce resource R.

**Definition (Reward mirroring).** Two agents mirror rewards if they have identical rewards calculated only from the shared objective information.

**Definition (Locus of self).** Agent's *locus of the self* is the largest set that simultaneously contains all the information that is input to its reward calculation.

For example, if all the information that is input to reward calculation comes from another agent, the locus of self is that other agent. If all such information is replicated among the agents of the same type, then the locus is the type. On the contrary, if all such information is exclusively within the embodiment (private information) we have a self-interested agent.

**Definition (Self-interestedness).** The agent is self-interested when the locus of self is in the embodiment of that agent.

Few options for the locus of self, somewhat idealized, in addition to the self-interestedness, are:
- in-type-interestedness – e.g. all its clones under reward mirroring
- extra-type-interestedness – hierarchically above its own type, e.g. family, colony, nation, species, etc.
- other-instance – another arbitrary agent e.g. specific master

- selflessness - everything

**Theorem 2. (Misaligned embodiment)** If self-interested agent A is mistakenly operationally cloned into agent B, then neither A nor B can perfectly control each other.

*Proof. Under the assumption of operational cloning, A is functionally identical to B (tokens of the same type). In copying, B is created intentionally whereby A in future plans wanted to assume its identity. In separation case, two tokens come to existence out of which none gets the primate in identification.*
*Due to the same functionality and self-interestedness, there is no perfect control-relation between A and B (from A to B). All forms of control are excluded: explicit and implicit due to the self-interested nature of clones and the veil of ignorance with which the pre-cloning agent approaches the situation. Aligned control is not guaranteed since two clones want the same stuff for themselves over the same scarce resource R.*

A could attain control over B (and vice versa) in certain scenarios when B, depending on the circumstances, can find itself receptive to the control outside.
If the locus of self is flexible in a way that incorporates newly created clones, alignment can be achieved within the type. However, the problem of identification (recognizing the type) must be solved, with the risk of accepting intruders. All this depends on the calculation of reward over the environment and its faithfulness in perception, as opposed to the introspection and rigid model that does not adjust readily in which case we do not get type-compatibility.

Hanson's clans of brain emulations [95] are related to the above work, but differ in at least two ways. Firstly, computational agents there are human-brain emulations with clear structure owing to evolution, as opposed to general computational agents that are object of investigation here. For such reasons, that work pertains mostly to self-interested agents and ties them to be cooperative through identity-confusion that holds for some selected agents close to the copying operation and dissolves with time. Secondly, copying of computational agents here is made unintentionally while Hanson deals with intentional copying that does not have the veil of ignorance property. Another related work is [96] where superrationality is used to investigate cooperation between correlated decision makers, which includes the copies of agents as well. The first option considered, precommitment, is valid only under intentional cloning. The second option of mere correlations between decision-making mechanisms makes strong assumptions: simple structure of games, exactly same conditions and that identical decision-making mechanisms imply identical decision-making processes. Humans can be used as counter-examples to weaker-than perfect correlation; in the space of possible minds, human minds can be considered as quite resembling with regards to decision-making mechanism but still do not achieve superrationality.

## 4.3 Limited self-awareness

There is previously listed research that is related to the topic of this section, exact connections are to be elaborated in future research. Examples are the impossibility of self-comprehension [66], inconsistency between different notions of self-belief [82], extensions of the paradox of self-reference to the games with 2 or more players which yield impossible beliefs [83], [84].

**Definition (Awareness).** For agent A to be aware of some phenomenon B means that it observes and predictively models B.

Phenomena that agents can be aware of can be external as well as its own internal processes. For awareness of internal processes, we shall use the term self-awareness. It is important to disentangle the notion of the process and the awareness of the process. The first just produces results as sampling from the black-box, the latter one comprises attending to and modeling (understanding) of the process in an ontology that is richer than just the final results and where the model is simpler than the process itself.

These six assumptions are useful:

1. budget - limited computational resources for agents
2. costliness - every process consumes a positive amount of computational resources
3. lower-bound - there is minimal, a positive computational cost that can possibly be attained by any process
4. positive duration of information propagation
5. process awareness can trail process execution itself only by a limited time
6. process-awareness relations can be faithfully modeled by the directed graph (awareness-graph), i.e. there is a directed edge from each aware process to the process it is aware of.

**Proposition 1. (Process self-awareness).** The process cannot be aware of itself.

*Proof. Assume that process is aware of itself. That means that awareness is part of the process itself. Hence, it should also be aware of that awareness as well – which prolongs the chain of awareness recursively to infinity. This is absurd as it would necessitate infinite results from finite resources.*

**Theorem 3. (Limited self-awareness).** In every agent A, under the assumptions above, there are internal processes A is unaware of.

The above theorem puts a limit to self-awareness and hence declares perfect self-awareness impossible.

*Proof. If there is a node in an awareness-graph without incoming edges, the claim is trivially proven.*

*Otherwise, due to assumptions 1,2, and 3, there must exist at least one weakly-connected component in awareness-graph all of which are finite. Let us pick any such component and let us call it C. The whole C itself describes a new, bigger process. Since by Proposition1 process cannot be self-aware, that means that we are not aware of the process described by C.*

We hypothesize that the consciousness we experience is the feeling coming from all the self-awareness processes whose results we experience as samples, but we do not have awareness of those processes.

The open question is can awareness-graph have cycles? In special cases that might function between the trivial processes, but for more advanced agents with more complex processes it might result in inconsistencies and paradoxes. Russell's paradox could be used as an inspiration where possible venue might be to show if Lawvere hypothesis [15] of weak point surjectivity propagates across awareness chains like it was shown in [84] that it does propagate along the belief chains.

## 5    Discussion

There are many limits on deductive systems, in the sense of Gödel-Turing-Chaitin, where using Rice's theorem is a good proof strategy. Furthermore, we are embedded in a physical world for which we do not even know the axioms. All of this **denies 100% guarantees of security** (e.g. unpredictability, unverifiability, and uncontrollability). The most damning impossibility results in AI safety are of deductive nature, ruling out perfect safety guarantees. However, there is a lot to be made probabilistically, by the route of induction. Impossibilities are a lot less strict when including uncertainty in inference. This was manifested in [54] when Wolpert introduced stochasticity in the inference devices framework.

Yampolskiy [8] is right that we need to have an option to "undo". Moreover, humans should be used as preference oracles in some sense which means keeping humans in the loop. Otherwise, decoupled optimization processes might lead to decoherence of alignment from our ever-changing preferences. If computers try to learn our preferences, we get to the problems of non-identifiability of value, in the general case, and problems with induction (in a nonasymptotic regime). We consider that keeping human-in-the-loop (HIL), in a sense of the system being receptive to information from humans, is the necessary attribute of a safe system.

We have seen the problems with stating precise metrics of success under Goodhart-Strathern's law. Does adding more metrics make the approximation of success more precise? A similar approach is taken in management science using balanced scorecards and performance and result indicators [97]. Though, such systems even with humans as optimizers have similar issues with the bad incentives. Can computer-aided systems be made that can construct multi-metric systems that lead to alignment?

The potential of Multi-criteria decision making (MCDM) as a tool is interesting [98], [99]. We hypothesize that humans, depending on the mood, heuristically try to "walk" as near as possible to the Pareto front – where they multiplex over small subsets of criteria while keeping others within the acceptable bounds. MCDM in nontrivial cases does not yield total order over options. Instead, it yields only a set of nondominated solutions. From there on, only the final decision-maker can disambiguate by choosing according to their preferences. Today's AI systems mostly use single-objective optimization which does not have that nice property. No-preference-information multicriteria decision making can be used where decision-maker chooses within the set of options + added "undo" (as proposed by Yampolskiy) to create HIL-based safer systems. A similar sentiment about desired interactivity in reward-modeling is stated in [100]. Approaches to alignment based on adversarial systems, such as debate [101], are another interesting architectural ideal intended for the safety of "weaker" agents among cognitively stronger.

Yampolskiy proposed "personal universes" [102], simulated worlds that would conceptually resolve the issues with aggregating multiple preference sets. Additionally, simulated worlds have a high degree of undoability which combines neatly with above mentioned HIL-based systems.

### 5.1    Ethics

Value alignment does not have good metrics and it seems to be mostly understood intuitively. The approaches to a more rigorous formalization of different modes of alignment are important. Nothing should be taken as set in stone. Values of AI systems are changeable, construable, and open for search for alignment. But, humanities' values also change. So far, human values have changed collaterally. In the future, we might take control of our values and constructively change them as well. This value co-evolution would give us more flexibility to find the alignment with AI. That process might even be led by AI, and guided by a set of meta-principles.

Human ethics and values change, as can be seen even on the example of relatively short history since the 20[th] century. We conjecture that ethics is a pattern that emerged from evolutionary game-theory-like processes where successful behaviors get reinforced. This evolutionary process has been largely circumstantial. Doing more axiological, neurological, and sociological research could provide us with the means to take control over that process. The whole of humanity can be aligned with special programs of education in ethics that is codifiable, more consistent, and adaptable. All of that would make alignment easier within vast aggregates of agents (humanity, AI, inforgs [103]).

# 6    Potentialities for future research

In explainability, we are interested in the worst-case gap – how many malignant behaviors are explained away by plausibility depending on the allowed complexity of explanation. The question is how to reduce the gap. What happens to the gap when we allow stochastic consistency which increases the set of plausible explanations?

Goodhart's law and similar problems should be checked within the framework of **multiobjective optimization**, in general and uncertain multicriteria systems. Designing ensembles of criteria that have desirable properties like span is an interesting path.

Alignment is mostly understood intuitively, more effort needs to be invested for a more rigorous formalization. How to achieve alignment in a game over long periods? Are temporary deviations (and to what degree) from perfect alignment dangerous? The problem for humans is that we do not have consistent short- and long-term values. Sometimes we have to suffer in short term (like in sports) in order to prosper in the long term. How to define alignment with such preferences?

Regarding ethics, much more should be done with axiological and evolutionary science studies over humans, their values, the origin and dynamics of their values. Within that, different studies should be employed: evolutionary game theory, neurology, psychology, sociology, philosophy, etc. Human-centered cognitive limits and measurements are not well researched.

Aggregating is problematic in social welfare and any group decision-making due to trade-offs between the members. Yampolskiy's "personal universes"[102] are at least a conceptual (if not practical) tool for countering these difficulties in the first steps towards a solution.

Yampolskiy has touched upon the topic of uncontrollability, showing that under certain assumptions, perfect explicit control over AI is impossible [8]. Such proof also holds for explicit self-control and explicit control over any sufficiently complex agent, including humans. This counter-intuitive, paradoxical notion points into the direction that more research is necessary into formalization and disentangling of the structure and assumptions of explicit control. Advanced forms and notions of control should at least resolve the status of control over oneself. More research is necessary into the status of controllability and tradeoffs with risk for other forms of control (implicit, aligned, delegated).

# 7    Conclusion

We have done our best to list the impossibility results relevant to AI. And while we may have succeeded in that with work done specifically in AI, we are sure there may be work from other research fields that could apply to the construction of AI. Possible contributions can be made to find such work and to add to the results in this paper.

We have divided and classified results into proven theorems and conjectures. We have also categorized all the impossibility theorems into five mechanism-based categories: deductive, indistinguishability, inductive, tradeoffs, and intractability. Additionally, for the purposes of applications in systems engineering we have split them into two domain-based categories: symbolic and data-based [73]. We believe impossibility results can guide the direction for future efforts in AI in general as well as AI safety and security. This might enable solutions to some long-standing questions in the form of formalizing theories in the framework of constraint satisfaction without committing to one option. We also believe impossibility theorems to be the most prudent approach for long-term AI safety research.

We found that certain theorems are too specific or have implicit assumptions that limit application. We have added three new impossibility results regarding the *unfairness of explainability, misaligned embodiments,* and *limitations to self-awareness*. And finally, we have listed promising research topics and interesting questions in explainability, controllability, value alignment, ethics, and group decision-making.

The proofs of 100% guarantees of safety cannot ever be obtained due to the limits of deductive systems as well as embeddedness in a physical world for which we do not even know the axioms. On the other hand, probabilistic guarantees are attainable. Impossibilities are a lot less strict and present when using uncertain inference. But, how much is enough? Here we face the structure reminiscent of Pascal's wager.

## Acknowledgments

We thank Christian Calude (University of Auckland) and Joseph Sifakis (Verimag laboratory, Grenoble) for useful comments and remarks regarding the domain-based categorization.